\documentclass{article}

\usepackage[nonatbib,final]{nips_2018}
\usepackage[numbers]{natbib}

\usepackage[utf8]{inputenc} 
\usepackage[T1]{fontenc}    
\usepackage{hyperref}       
\usepackage{url}            
\usepackage{booktabs}       
\usepackage{amsfonts}       
\usepackage{nicefrac}       
\usepackage{microtype}      
\usepackage{enumitem}
\usepackage{subfig}

\usepackage{bbm}
\usepackage{amsmath}
\usepackage{graphicx}

\title{Model-Based Reinforcement Learning \\ for Sepsis Treatment}

\author{Aniruddh Raghu\\
MIT\\
Cambridge, MA \\
\texttt{araghu@mit.edu}
\And 
Matthieu Komorowski
\\Imperial College
\\London, United Kingdom\\
\texttt{matthieu.komorowski@gmail.com}
\And 
Sumeetpal Singh
\\Cambridge University
\\Cambridge, United Kingdom\\
\texttt{sss40@cam.ac.uk}
}

\begin{document}

\maketitle

\begin{abstract}
Sepsis is a dangerous condition that is a leading cause of patient mortality. Treating sepsis is highly challenging, because individual patients respond very differently to medical interventions and there is no universally agreed-upon treatment for sepsis. 
In this work, we explore the use of continuous state-space model-based reinforcement learning (RL) to discover high-quality treatment policies for sepsis patients. Our quantitative evaluation reveals that by blending the treatment strategy discovered with RL with what clinicians follow, we can obtain improved policies, potentially allowing for better medical treatment for sepsis.
\end{abstract}

\section{Introduction} 
Sepsis (severe infections with organ failure) is a dangerous condition that is a leading causing of patient mortality and is expensive for hospitals to treat \citep{sepsismortality,sepsiscost}. 
Sepsis is often managed by giving patients intravenous fluids and vasopressors. Different dosage strategies for these two interventions can greatly affect patient outcomes, which demonstrates how important treatment decisions are \citep{waechter2014interaction}. However, clinicians still lack decision support tools to assist them when treating patients with sepsis~\citep{rhodes2017surviving}. This work explores the use of reinforcement learning (RL) to provide medical decision support for sepsis treatment, developing on earlier research \citep{komorowski, raghu2017deep}. We consider the use of continuous state-space model-based RL algorithms and observational data from intensive care units (ICUs) to deduce medical treatment strategies (or \emph{policies}) for sepsis. We demonstrate how blending our learned policies with what clinicians follow could lead to better medical treatment for sepsis.

\section{Preliminaries}
\paragraph{Reinforcement Learning (RL):} In the RL problem, an agent's interaction with an environment can be represented by a Markov Decision Process (MDP).
At every timestep $t$, an agent observes the current state of the environment $s_t$, takes an action $a_t$ according to a \emph{policy}, $\pi(a_t|s_t)$, receives a reward $r_t$, and transitions to a new state $s_{t+1}$ with probability $P(s_{t+1}|s_t,a_t)$, where $P$ is the transition probability distribution. The \emph{return} is defined as $R_{0:T-1} = \sum_{t=0}^{T-1} \gamma^{t}r_{t}$, where $\gamma$ captures the tradeoff between immediate and future rewards, and $T$ is the terminal timestep. The value of a policy $\pi$, $V^\pi$, represents its quality, and is defined as the expected return over trajectories $\{H_i\}$ it generates: $V^\pi=\mathbb{E}_{H\sim P^\pi_H}\big[R_{0:T-1}(H)\big]$. 
We consider continuous state-space model-based RL, where we model the transition distribution $P$ and then using this to find a good-quality policy $\pi$. Prior work has considered model-free RL \citep{raghu2017deep}, and discrete state-space model-based RL \citep{komorowski}.

\paragraph{Related work:} RL has previously been used in healthcare settings. Sepsis treatment strategy was tackled by \cite{komorowski}, where model-based RL with discretized state and action-spaces were used to deduce treatment policies. \cite{raghu2017deep} extended this work to consider continuous state-space model-free RL. In our work, we build on both approaches: we explore the use of continuous state-space model-based RL, which is challenging due to having to model the transition distribution of the MDP, but also potentially more data-efficient. We also extend our quantitative evaluation based on recent work \citep{raghu2018behaviour}.

\paragraph{Data and preprocessing:} In this work, we consider a cohort of patients from the Medical Information Mart for Intensive Care (MIMIC-III v1.4) database  \citep{mimic} fulfilling the Sepsis-3 criteria \citep{sepsis3} (17,898 in total), as in \cite{raghu2017deep}. For each patient, we have relevant physiological parameters including demographics, lab values, vital signs, and intake/output events, discretized into 4 hour timesteps. In order to capture temporal patterns in a patient's physiology, the state representation used in this work concatenates the previous three timesteps' raw state information to the current time's state vector, resulting in a vector of length 198, which is the state $s_t$ in the underlying MDP. See Section \ref{sec:appendix_features} for the full list of features used at each timestep.

\paragraph{Actions and rewards:} We define a $5\times5$ action space for the medical interventions, as in \cite{raghu2017deep} covering dosages of intravenous (IV) fluid (volume adjusted for fluid tonicity) and maximum vasopressor (VP) dosage in a given 4 hour window. More information is provided in Section  \ref{sec:appendix_action_space}.
The reward function follows from \cite{raghu2017deep}, and is clinically guided: positive rewards are issued at intermediate timesteps for improvements in a patient's wellbeing (with improvement being defined by reductions in severity scores, such as SOFA), and negative rewards for deterioration. At the terminal timestep of a patient's trajectory, a reward is assigned that is positive in the case of survival, and negative otherwise. See Section \ref{sec:appendix_reward} for more information.

\section{Model-based RL}
We now explore our model-based RL approach to learning sepsis treatment strategies.

\subsection{Environment modelling}
We first construct a model to represent the transition dynamics of the underlying MDP, which we address by learning an environment model. This framing of the problem is motivated by the discussion in Nagabandi et al.~\cite{nagabandi2017neural}. 
We consider the task of predicting $\Delta_t = s_{t+1} - s_t$, the change in a patient's physiological state, which is achieved by learning a function $f(h_t;\theta)$ where:
${\Delta}_{t} =  f(h_t;\theta) + \mathbf{\epsilon}; \quad  \mathbf{\epsilon} \sim \mathcal{N}(\mathbf{0},\mathbf{I})$, and 
$h_t = g\left(s_t,a_t, s_{t-1},a_{t-1}, \dots   \right)$.

The function $g$ concatenates the current state-action pair $(s_t, a_t)$ with the previous three timesteps' state-action pairs. This is to capture some amount of historical information about the patient's physiology. Cross-validation performance (in terms of prediction mean squared error) motivated the use of three past timesteps of information.

We considered several different environment models:
\begin{itemize}[nosep,leftmargin=*]
    \item Linear regression to predict ${\Delta}_{t}$
    \item Neural network: the best architecture had two fully connected layers with ReLU activations, and batch normalization \citep{batchnorm} after each layer. This was trained using Adam \citep{adam}.
    \item Recurrent neural network (RNN); specifically, an LSTM \citep{hochreiter1997long}. This was trained using Adam \citep{adam}.
    \item Bayesian neural network (BNN): this is advantageous as it can represent the full predictive distribution over $\widehat{\Delta}_t$ instead of providing point estimates. We based our BNN modelling approach on that of \cite{depeweg2016learning}. The best BNN had two hidden layers, each with 32 hidden units and tanh nonlinearities, and was implemented using the Autograd library \citep{autograd}. A full description of BNN modelling is provided in Section \ref{sec:BNN_appendix}.
\end{itemize}

 
\paragraph{Results:} Table \ref{tab:modelbased_mse} shows the mean squared error (MSE) when predicting $\Delta_t$ when using different environment models: Linear Regression (LR), feedforward Neural Network (NN), Recurrent Neural Network (RNN), and Bayesian Neural Network (BNN).

\begin{table}[h]
\centering
\caption{Mean squared error (MSE) on a held-out validation set when predicting $\Delta_t$ for different environment models: Linear Regression (LR), feedforward Neural Network (NN), Recurrent Neural Network (RNN), and Bayesian Neural Network (BNN).}
\label{tab:modelbased_mse}
\begin{tabular}{@{}lcccc@{}}
\toprule
    & LR    & NN    & RNN            & BNN   \\ \midrule
MSE & 0.195 & 0.171 & \textbf{0.122} & 0.220 \\ \bottomrule
\end{tabular}
\end{table}

The RNN obtains the best MSE by this metric. Analysis reveals that this is because the predictions produced by the RNN at large timesteps are very accurate. 
However, the performance of the RNN at small values of $t$ are poor; the RNN does not have the capacity to predict $\Delta_t$ accurately at small $t$. The feedforward neural network is therefore preferred as an environment model.

\begin{figure}[htbp]
\centerline{     
\subfloat[Rollout 1]{\label{fig:mbrl_traj_1}\includegraphics[width=55mm]{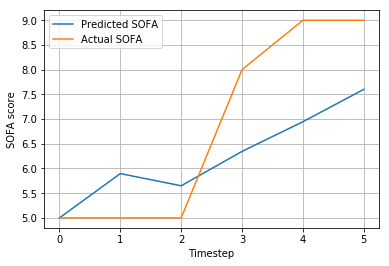}}
\subfloat[Rollout 2]{\label{fig:mbrl_traj_2}\includegraphics[width=55mm]{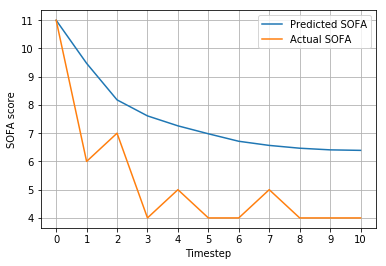}}
\subfloat[Rollout 3]{\label{fig:mbrl_traj_6}\includegraphics[width=55mm]{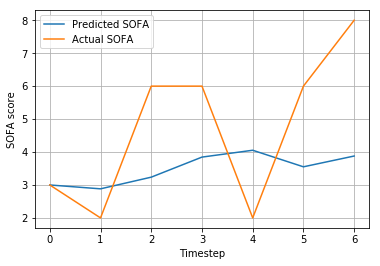}}}
\caption{Rollouts from the neural network environment model for the SOFA score, an important physiological feature. Predictions from the network can approximately capture the trend in SOFA score in certain cases, but not others.}
\label{fig:mbrl_traj}
\end{figure}

Figure \ref{fig:mbrl_traj} shows some examples of rollouts for the SOFA score using the neural network as an environment model, using the first state in a trajectory and clinician actions at subsequent timesteps.
The neural network  can sometimes represent the overall trend in SOFA score over a patient's trajectory accurately; this is most noticeable for Figures \ref{fig:mbrl_traj_1}, \ref{fig:mbrl_traj_2}. This task is however very challenging -- there are large changes in the value of the SOFA score at certain timesteps, and there are timesteps where the values do not change at all. The accuracy of the predictions worsens as we increase the length of the rollout. The modelling performance is not perfect, but given that the model can capture the overall SOFA trend, it may be sufficiently accurate to aid in discovering improved policies. However, further work is necessary on this task.

\subsection{Policy search}
\label{sec:policysearch}
After learning an environment model, policy search algorithms are used to find an appropriate policy $\pi$.
It is important not to stray too far from the policy used by clinicians for interpretability reasons; therefore, the approach to discover improved treatment strategies proceeded as follows:

\begin{itemize}[nosep,leftmargin=*]
\item Learn the parameters $\phi$ describing the clinician policy, $\mu(a|h_t;\phi)$, using supervised learning.
\item Initialise the parameters $\theta$ of a new policy $\pi(a|h_t;\theta)$ using $\phi$.
\item Using the learned environment model, simulate rollouts from $\pi(a|h_t;\theta)$. Use a policy improvement algorithm with a small learning rate (to stay close to $\mu$) to improve $\pi$.
\end{itemize}

\paragraph{Modelling clinician behaviour:} We formulated learning the clinician policy $\mu$ as a behaviour cloning task, where we learned parameters $\phi^{\ast}$ that minimised the average cross-entropy between the predictive distribution over actions from the neural network and the label distribution (encoding the action taken by the clinician). 
The best-performing model was a two-layer feedforward neural network with 64 hidden units in each layer, with ReLU activations. An L2 regularisation penalty was added to the loss to prevent overfitting.

\paragraph{Policy improvement:} To improve $\pi$, we compare the performance of two policy improvement algorithms: the REINFORCE Policy Gradient algorithm  (PG) \citep{williams1992simple}, and the Proximal Policy Optimization algorithm (PPO) \citep{ppo}. Our procedure firstly involves sampling a random trajectory from the dataset, to get a starting state. We then perform a rollout from this starting state for 10 timesteps, using actions sampled from $\pi(a|h_t;\theta)$. We use the policy improvement algorithms with these trajectories to improve $\pi$.

\subsection{Evaluation of learned policies}
\label{sec:evalmbrl}
Qualitative analysis of learned treatment policies (as seen in previous work by \cite{raghu2017continuous}) does not necessarily provide an absolute indicator of performance, explored in more detail by \cite{gottesman2018evaluating}. We therefore consider quantitative evaluation of learned policies, using Off-Policy Policy Evaluation (OPE). Our evaluation uses the Per-Horizon Weighted Importance Sampling (PHWIS), Per-Horizon Weighted Doubly Robust (PHWDR), and Approximate Model (AM) estimators, discussed in \cite{raghu2018behaviour}. We follow the suggestions of \cite{raghu2018behaviour} when constructing these estimators.  To model the clinical policy (behaviour policy) for these estimators, an approximate k-Nearest Neighbours (kNN) model with $k=250$ was used. To obtain the AM estimator and the AM terms for the PHWDR estimator, the Fitted Q Iteration (FQI) algorithm \citep{ernst2005tree} was used with a random forest with 80 trees. We do not use recent model-based estimators, introduced by \cite{liu2018representation}, because of the challenge in learning full environment models in this domain.

We present results here for the PPO algorithm; the PG algorithm performed relatively poorly in comparison. We also consider blending the clinician's policy and the RL algorithm's policy in different severity regimes (deduced by SOFA score); the environment model is not as accurate in high and low severity regimes (due to high stochasticity and less signal respectively). Given the safety-critical nature of this domain, we rely on the clinician's policy in regimes where the model has inferior performance. For comparison, note that the value of the clinician's policy, $V^{\pi_b}$, obtained by averaging returns in the dataset, is found to be 9.90. The policy evaluated is denoted $\pi_e$.

Table \ref{tab:phwis_stratify} presents results using the PHWIS, PHWDR, and AM estimators.
These results show that the AM estimator does not appear to discriminate significantly between the learned policies; all policies have similar values.
When considering the PHWIS and PHWDR estimators, it appears that the best policy is obtained when relying on the clinical policy in the low and high severity regimes, and the RL-learned policy in the medium severity regime. The two estimation procedures show reasonable agreement in estimated value, though there are discrepancies between them. However, as both show similar trends, it may be possible to conclude that this policy does improve on what clinicians currently follow. Investigating the specific policy discovered in this regime could offer valuable clinical insight.
The result obtained can also be interpreted --  in low/high severity regimes, clinicians may have a set protocol they follow, which performs well; it therefore does not make sense to deviate from this. In the medium severity regime, there is more variability in clinician action, and model-based RL is effective in identifying suitable courses of action. 

\begin{table}[h]
\caption{OPE results using the PHWIS, PHWDR, and AM estimators when relying on combinations of the clinician's policy and the PPO policy. Results are also shown for following the clinical policy in all regimes. By relying on the clinician policy in regimes where the environment model is weaker, improvements in performance are observed for the PHWIS and PHWDR estimators.}
\centering
\begin{tabular}{@{}lllccc@{}}
\toprule
Low SOFA  & Medium SOFA & High SOFA & $\hat{V}^{\pi_e}_{\text{PHWIS}}$ & $\hat{V}^{\pi_e}_{\text{PHWDR}}$ & $\hat{V}^{\pi_e}_{\text{AM}}$ \\ \midrule
PPO       & PPO         & PPO       & 10.8            & 11.5        &  9.33    \\ 
PPO       & PPO         & Clinician & 11.7            & 11.8        &  9.35     \\ 
Clinician & PPO         & Clinician & \textbf{12.1}   & \textbf{12.8} &  9.35     \\ 
PPO       & Clinician   & PPO       & 7.63            & 7.58        &  9.34      \\ 
Clinician & Clinician   & Clinician & 10.2            & 9.87        &  \textbf{9.36}     \\ \bottomrule
\end{tabular}
\label{tab:phwis_stratify}
\end{table}

\section{Conclusions}
In this work, we proposed a method to learn medical treatment strategies for sepsis with model-based reinforcement learning. We used a neural network to model physiological dynamics, and demonstrated how to learn treatment policies using a behavioural cloning objective and policy improvement algorithms (Proximal Policy Optimisation). Our quantitative evaluation revealed we could find potentially improved treatment strategies by blending clinical and RL-learned policies, which could lead to improved treatment for sepsis patients.

\section{Acknowledgements}
The authors would like to thank Omer Gottesman, Yao Liu, Finale Doshi-Velez, and Emma Brunskill for helpful comments and suggestions.

\bibliography{nips}
\bibliographystyle{plain}
\clearpage
\section{Appendix}
\subsection{Cohort definition}

Following the latest guidelines, sepsis was defined as a suspected infection (prescription of antibiotics and sampling of bodily fluids for microbiological culture) combined with evidence of organ dysfunction, defined by a Sequential Organ Failure Assessment (SOFA) score greater or equal to 2 \citep{sepsis3}. We assumed a baseline SOFA of zero for all patients. For cohort definition, we respected the temporal criteria for diagnosis of sepsis: when the microbiological sampling occurred first, the antibiotic must have been administered within 72 hours, and when the antibiotic was given first, the microbiological sample must have been collected within 24 hours \citep{sepsis3}. The earliest event defined the onset of sepsis. We excluded patients who received no intravenous fluid, and those with missing data for 8 or more out of the 48 variables. This method yield a cohort of 17,898 patients.

The resulting cohort is described in Table \ref{tab:cohort}.
\begin{table}[htbp]
\centering
 \caption{Summary statistics for the patient cohort.} 
  \label{tab:cohort} 
\begin{tabular}{@{}lllll@{}}
\toprule
              & \% Female & Mean Age & Hours in ICU & Total Population \\ \midrule
Survivors     & 43.6      & 63.4     & 57.6         & 15,583           \\
Non-survivors & 47.0      & 69.9     & 58.8         & 2,315            \\ \bottomrule
\end{tabular}
\end{table}
\subsection{Data extraction}

MIMIC-III v1.4 was queried using pgAdmin 4. Raw data were extracted for all 47 features and processed in Matlab (version 2016b). Data were included from up to 24 hours preceding the diagnosis of sepsis and until 48 hours following the onset of sepsis, in order to capture the early phase of its management including initial resuscitation, which is the time period of interest. The features were converted into multidimensional time series with a time resolution of 4 hours. The outcome of interest was in-hospital mortality.

\subsection{Model Features}
\label{sec:appendix_features}

\textbf{Choice of features}: the included features were chosen to represent the most important parameters clinicians would examine when deciding treatment and dosage for patients; however, some factors not included in our feature vector could serve as confounding factors. Exploring the effect of these is an important future direction. 

The physiological features used in our model are:

\textbf{Demographics/Static}: Shock Index, Elixhauser, SIRS, Gender, Re-admission, GCS - Glasgow Coma Scale, SOFA - Sequential Organ Failure Assessment, Age \newline \newline
\textbf{Lab Values}: Albumin, Arterial pH, Calcium, Glucose, Hemoglobin, Magnesium, PTT - Partial Thromboplastin Time, Potassium, SGPT - Serum Glutamic-Pyruvic Transaminase, Arterial Blood Gas, BUN - Blood Urea Nitrogen, Chloride, Bicarbonate, INR - International Normalized Ratio, Sodium, Arterial Lactate, CO2, Creatinine, Ionised Calcium, PT - Prothrombin Time, Platelets Count, SGOT - Serum Glutamic-Oxaloacetic Transaminase, Total bilirubin, White Blood Cell Count \newline \newline
\textbf{Vital Signs}: Diastolic Blood Pressure, Systolic Blood Pressure, Mean Blood Pressure, PaCO2, PaO2, FiO2, PaO/FiO2 ratio, Respiratory Rate, Temperature (Celsius), Weight (kg), Heart Rate, SpO2 \newline \newline
\textbf{Intake and Output Events}: Fluid Output - 4 hourly period, Total Fluid Output, Mechanical Ventilation \newline \newline
\textbf{Miscellaneous}: Timestep 

\subsection{Action space}
\label{sec:appendix_action_space}
We discretized the action space into per-drug quartiles based on all non-zero dosages of the two drugs, and converted each drug at every timestep into an integer representing its quartile bin. We included a special case of no drug given as bin 0. This created an action representation of interventions as tuples of (total IV in, max VP in) at each time. We choose to focus on this action space given the uncertainty in clinical literature of how to adjust these interventions on a per-patient basis, and also their crucial impact on a patient's eventual outcome \citep{waechter2014interaction}.

\subsection{Reward function}
\label{sec:appendix_reward}
As our objective is to improve patient survival, the model should receive a reward when the patient's state improves, and a penalty when the patient's state deteriorates. This reward should be comprised of the best indicators of patient health; in this situation, these indicators include the patient's SOFA score (which summarizes the extent of a patient's organ failure and thus acts as a proxy for patient health) as well as the patient's lactate levels (a measure of cell-hypoxia that is higher in septic patients because sepsis-induced low blood pressure reduces oxygen perfusion into tissue). An effective reward function should penalize high SOFA scores as well as increases in SOFA score and reward decreases in SOFA scores between states. Similarly, for lactate, increases in lactate should be penalized while decreases in lactate should be rewarded. 

We opted for a reward function for intermediate timesteps as follows:
$$r(s_{\textit{t}}, s_{\textit{t+1}}) = C_{0}\mathbbm{1}(s^{\textnormal{SOFA}}_{\textit{t+1}} = s^{\textnormal{SOFA}}_{\textit{t}}\ \& \  s^{\textnormal{SOFA}}_{\textit{t+1}} > 0) + C_{1}(s^{\textnormal{SOFA}}_{\textit{t+1}}-s^{\textnormal{SOFA}}_{\textit{t}}) + C_{2}\tanh(s^{\textnormal{Lactate}}_{\textit{t+1}} - s^{\textnormal{Lactate}}_{\textit{t}})$$
We experimented with multiple parameters and opted to use $C_{0} = -0.025$, $C_{1} = -0.125$, $C_{2} = -2$. 

At terminal timesteps, we issue a reward of $+15$ if a patient survived their ICU stay, and a negative reward of $-15$ if they did not.

The reason behind the chosen parameters for the intermediate reward, as well as the form of the equation above, was to ensure that the reward was limited in magnitude and would not eclipse the terminal timestep reward at the end of each patient's trajectory. This was also the motivation behind using the $\tanh$ function when dealing with lactate changes: because the maximum change was significantly higher than the average change between timesteps, we opted to use a $\tanh$ to cap the maximum reward/penalty to $|C_{2}|$.




\subsection{BNN modelling}
\label{sec:BNN_appendix}
These models represent the full predictive distribution over $\widehat{\Delta}_t$ instead of providing point estimates. 

The description provided here is based on that in Depeweg et al.~\citep{depeweg2016learning}. 

As with non-Bayesian neural networks, a likelihood model is specified, mapping the input to the output, using a series of linear transformations and pointwise nonlinearities. 
A prior distribution is specified over the values of the parameters of the weight matrices and bias vectors; in this case, it is a Gaussian with zero mean and specifiable variance: $\theta_i \sim \mathcal{N}(0, \sigma_i^2)$.

In BNN modelling, we aim to form the posterior distribution over the parameters given the training data, consisting of $(h_t, \Delta_t)$ pairs, and marginalise out the parameters to get the predictive distribution over $\widehat{\Delta}_t$. 

However, as this marginalisation procedure is intractable, we form a variational approximation to the posterior distribution, $q(\theta | \mathcal{D})$, and minimise the alpha-divergence between the true posterior and the variational approximation \citep{hernandez2016black}. This minimisation procedure aims to find the parameters of the approximate posterior; that is, the mean and variance of the Gaussian distributions that make up the approximate posterior distribution.

Samples from the predictive distribution are then obtained by sampling parameter values from the approximate posterior and averaging: 
\begin{equation*}
\begin{split}
p(\widehat{\Delta}_t | h_t, \mathcal{D}) &= \int p(\widehat{\Delta}_t | h_t, \mathcal{D}, \theta) q(\theta | \mathcal{D}) d\theta \\
 & \approx \frac{1}{N}\sum_{i=1}^{N} p(\widehat{\Delta}_t | h_t, \mathcal{D}, \theta_i); \quad \theta_i \sim q(\theta | \mathcal{D})
\end{split}
\end{equation*}

The best BNN had two hidden layers, each with 32 hidden units and tanh nonlinearities. It was implemented using the Autograd library \citep{autograd}.

\end{document}